# SharpGAN: Receptive Field Block Net for Dynamic Scene Deblurring

Hui Feng, Jundong Guo, Sam shuzhi Ge

**Abstract**

When sailing at sea, the smart ship will inevitably produce swaying motion due to the action of wind, wave and current, which makes the image collected by the visual sensor appear motion blur. This will have an adverse effect on the object detection algorithm based on the vision sensor, thereby affect the navigation safety of the smart ship. In order to remove the motion blur in the images during the navigation of the smart ship, we propose SharpGAN, a new image deblurring method based on the generative adversarial network. First of all, the Receptive Field Block Net (RFBNet) is introduced to the deblurring network to strengthen the network's ability to extract the features of blurred image. Secondly, we propose a feature loss that combines different levels of image features to guide the network to perform higher-quality deblurring and improve the feature similarity between the restored images and the sharp image. Finally, we propose to use the lightweight RFB-s module to improve the real-time performance of deblurring network. Compared with the existing deblurring methods on large-scale real sea image datasets and large-scale deblurring datasets, the proposed method not only has better deblurring performance in visual perception and quantitative criteria, but also has higher deblurring efficiency.

## 1. Introduction

Smart ships are usually equipped with a variety of sensor devices to perceive their own motion state and the surrounding environment. Among them, the visual perception based on visible light is the core of smart ship perception technology. However, since the ship will inevitably sway during the navigation, motion blur will appear on the images collected by the vision sensor, which will adversely affect the object detection of obstacles, thereby reducing the safety performance of smart ship. Therefore, it is of great significance to remove motion blur from the images.

Due to the ill-posedness of the image deblurring problem, the traditional deblurring methods (Richardson, 1972; Lucy, 1974; Fergus et al., 2006; Shan et al., 2008; Krishnan and Fergus, 2009; Whyte et al., 2010; Xu et al., 2013) set up some constraints (such as uniform blur/non-uniform blur) and combine the prior information of the blurred images to establish a corresponding mathematical model to solve the latent sharp images. These methods have very high computational complexity, and the deblurring effect of the models also has great dependence on the setting of model parameters. In addition, since the establishment of the models relies on the corresponding constraints, these methods are difficult to adapt to the actual deblurring problem.

With the rapid development of deep learning in recent years, many image deblurring methods based on convolutional neural networks and generative adversarial networks have been proposed. Most of these methods (Noroozi et al., 2017; Nah et al., 2017; Gong et al., 2017; Kupyn et al., 2018) improve the network's ability to extract image features by stacking a large number of convolutional layers or ResBlocks (He et al., 2016) while obtaining a larger range of receptive fields to improve the quality of restored images. These methods have achieved good deblurring effects, but the ability of simply stacked convolutional layers to extract image features is not adequate. Secondly, although it is important to improve the clarity of blurred images, it is also necessary to ensure that restored images and sharp images have highly similar image features. The existing deep learning-based image deblurring methods (Kupyn et al., 2018) use the conv_3-3 layer of the VGG19 network (Simonyan and Zisserman, 2015) to construct the content loss function to improve the feature similarity between the restored images and the sharp images, but this does not make full use of the powerful image feature extraction ability of VGG19 network. In addition, in the actual navigation of the ship, the good real-time performance of the deblurring algorithm is the guarantee for the smart ship to make timely and correct decisions in the effective time, but the real-time performance of the existing deblurring algorithm is still not ideal.

In this paper, we propose SharpGAN in response to the above mentioned problems. The contribution can be summarized as follows: (*i*) In order to enhance the network's ability to extract image features and maintain better real-time performance to adapt to the actual intelligent navigation of ships, we introduce the RFBNet-s module into the deblurring network for the first time. It combines the advantages of inception and dilated convolution as well as lightweight structure, and is able to effectively extract image features by simulating the perceptual characteristics of human vision. (*ii*) To make the restored images and the sharp images have more similar image features, we proposed a new feature loss based that combines different levels of image features. The feature loss is constructed using the feature maps of multiple convolutional layers of the CNN, which makes full use of the powerful image feature extraction abilities of the CNN and can better guide the deblurring network to learn deblurring. (*iii*) The experiments conducted on both GOPRO and SMD datasets reveal that the SharpGAN can restore clearer blurred images than the existing algorithms and maintain the best prediction efficiency. In addition, it can also significantly improve the object detection performance of the real sea image.

## 2. Related Work

### 2.1 Image Deblurring

The process of image blurring can be represented by the following model:

$$I_B = K \bullet I_S + N \qquad (1)$$

where $I_B$ is the blurred image, $I_S$ is the latent sharp image, $K$ is the blur kernel, and $N$ is the noise. The latent sharp image $I_S$ is convolved with the blur kernel and noise is added to produce the blurred image $I_B$. According to whether the blur kernel is known or not, the method of image deblurring can be divided into non-blind deblurring method and blind deblurring method.

Most of the earlier image deblurring methods are non-blind image deblurring methods. These methods need to deconvolve the blurred image under the condition that the blur kernel is known to solve the restored image, such as Lucy-Richardson algorithm (Richardson, 1972; Lucy, 1974), Wiener filter algorithm (Helstrom, 1967) and algorithm based on total variational model(Rudin et al., 1992).Compared with the non-blind deblurring method, the blind deblurring method can restore the blurred images under the condition that the blur kernel is unknown. Since Fergus et al. (2006) successfully used the variational Bayesian method to achieve blind deblurring of images, many blind deblurring methods had been proposed one after another. Shan et al. (2008) used a piecewise function to fit the gradient distribution of the blurred image, and then used alternate iteration method to estimate the latent sharp image and blur kernel. Krishnan and Fergus (2009) employed the normalized sparsity measure to estimate the blur kernel, and the application of the regularization term based on the image gradient had improved the restoration effect of the blurred images. Xu et al. (2013) proposed the method based on the unnatural L0 norm to estimate the blur kernel. The convolutional neural network and Markov theory were used by Sun et al. (2015) to estimate the blur kernel at the patch level to remove non-uniform motion blur. The fully convolutional deep neural network was employed by Gong et al. (2017) to estimate the motion flow of the blurred images, and then the method of (Zoran and Weiss, 2011) is used to solve latent sharp images. Yan et al. (2017) adopted the half quadratic splitting algorithm to iteratively solve the latent sharp image and blur kernel based on the light and dark channel theory.These deblurring methods based on blur kernel theory and motion flow have certain deblurring effects, but the deblurring process of each image takes a lot of time.

With the rapid development of deep learning in recent years, many efficient end-to-end image deblurring methods had been proposed. Noroozi et al. (2017) added the skip connection to the deblurring network, so that the network only needed to learn the residual between the blurred image and the sharp image, thereby reducing the difficulty of network learning to deblur. Nah et al. (2017) used a large number of ResBlocks (He et al., 2016) to construct a multi-scale convolutional neural network, which gradually deblurred from low-resolution image to final high-resolution image. Based on the generation adversarial network (Goodfellow et al. 2014), Kupyn et al. (2018) proposed an end-to-end image deblurring network (DeblurGAN), which used the PatchGAN (Isola et al., 2017) as the discriminator and the network was optimized by adversarial loss and content loss. Since then, Kupyn et al. (2019) proposed a new deblurring method (DeblurGANv2) based on conditional generation confrontation network (C-GAN). DeblurGANv2 employed inceptionv2 as the backbone of the generator, and combined the feature pyramid (FPN) to assist the network in extracting blurred image features.

*2.2 Generative Adversarial Networks*

Inspired by the idea of two-person zero-sum game in game theory, Goodfellow et al. (2014) proposed the Generative Adversarial Network (GAN), which mainly includes two architectures: the generator G and the discriminator D. Among them, the generator uses noise data z (z obeys Gaussian distribution or other prior probability distributions) to generate real samples as much as possible to deceive the discriminator, that is, to continuously learn the potential data distribution in the real samples. The discriminator continuously discriminate the generated samples and the real samples to improve its discrimination ability, and the results of the discrimination are fed back to the generator to make it generate more realistic samples. The training process of GAN is a game process between the generator *G* and the discriminator *D*, and the loss function of the network can be expressed as:

$$\min_G \max_D V(D,G) = E_{x \sim P_{data}}[\log(D(x))] + E_{x \sim P_G(x)}[\log(1-D(x))] \quad (2)$$

where $P_{data}$ is the data distribution of the real samples, and $P_G$ is the data distribution of the generated samples.

The proposal of GAN is extremely imaginative and creative, but its original network structure is not perfect, and there are a series of problems such as mode collapse and training difficulty. The root of these problems is that the loss function of the GAN is to minimize the Jensen–Shannon (JS) divergence between the real samples and the generated samples. In the high-dimensional space, the JS divergence is often difficult to reflect the difference between the two data distributions, so it cannot provide effective gradient information for the network. In response to this problem, Arjovsky et al. (2017) proposed WGAN, which used Wasserstein distance to replace the JS divergence in the original GAN loss function and effectively reflected the difference between the real samples data

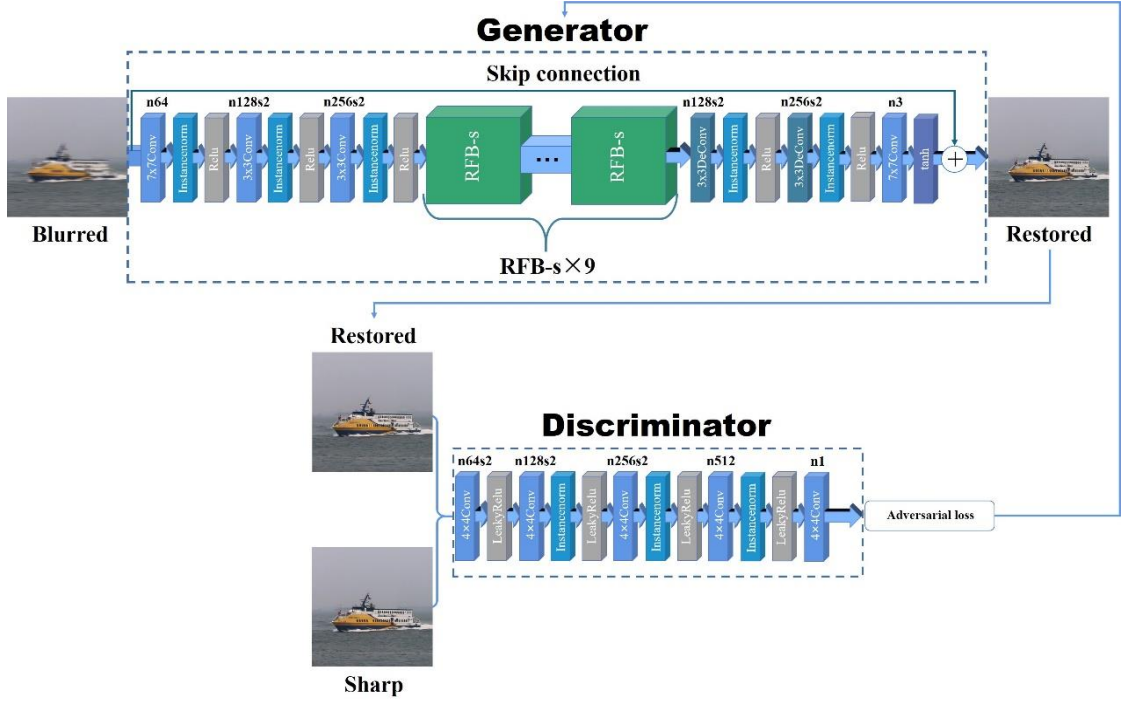

Figure 1: The architecture of the SharpGAN network. The generator of SharpGAN has 4 convolution layers, 2 deconvolution layers and 9 RFB-s modules, and the discriminator is similar to PatchGAN (Isola et al., 2017).

distribution and the generated samples data distribution. The loss function of WGAN was expressed as follow:

$$\min_{G} \max_{D \in D^*} V(D,G) = E_{x \sim p_{data}}[D(x)] - E_{x \sim p_G(x)}[D(x)] \quad (3)$$

where $D^*$ is the set of 1- Lipschitz functions. In order to make the discriminator meet the 1-Lipschitz continuity condition, Arjovsky et al. (2017) employed the weight clipping method to limit the weight of the discriminator to the range of [-c,c], but the selection of c can easily lead gradient vanishing/ exploding problem. For this reason, Gulrajani et al. (2017) proposed WGAN-GP, which added a gradient penalty term

$$E_x[\|\nabla_x D(x)\| - 1]^2 \quad (4)$$

to the loss function of WGAN. This made the training of GAN more stable and effectively solved the gradient vanishing/exploding problem of WGAN.

## 3. Proposed Method

In order to achieve end-to-end efficient motion deblurring, we propose the SharpGAN based on the generative adversarial network, which includes two parts: generator and discriminator. Given a blurred image, the generator is able to directly generate the restored image, and the discriminator guides the generator to better learn motion deblurring by distinguishing the sharp image from the restored image. The architecture of the SharpGAN network is shown in Figure 1.

### 3.1 Network Architecture

#### 3.1.1 Generator

In the proposed SharpGAN, the generator uses the blurred images as the input to generate the latent sharp images, as in (Kupyn et al., 2018). The front end of generator uses 3 convolutions to initially extract the blurred image features, and the back end of generator uses 2 deconvolution layers and 1 convolution layer to reconstruct the restored image with the same resolution as the input. The main difference with (Kupyn et al., 2018) is that in the middlepart of the generator, the ResBlocks is replaced by 9 RFBNet modules (Liu et al., 2018), which makes the generator more powerful in image feature extraction. RFBNet has two structures (Liu et al., 2018): RFB and RFB-s. In order to make generator more lightweight and improve the real-time performance of deblurring, we choose the RFB-s module with fewer parameters as the part of the generator, and its network structure is shown in Figure 2.

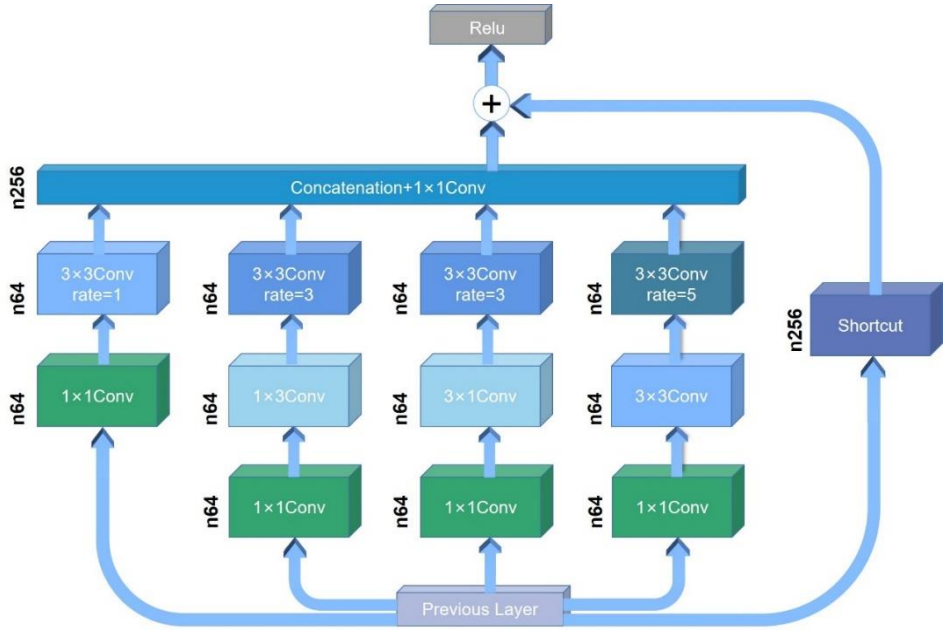

Figure 2: The architecture of the RFB-s module.

The RFB-s module can be divided into 5 branches. Each branch first uses a 1 × 1 conv-layer to reduce the dimension of the input data, so the network requires less calculation. Motivated by ResNet's skip connection (He et al., 2016), a branch called Shortcut is directly connected to the activation layer of the module, meanwhile, inspired by the network structure of Inception (Szegedy et al., 2015, 2016, 2017), convolution kernels of different sizes are set in the other branches of the module. The advantages are that different sizes of convolution kernels can obtain different receptive fields, which are able to enhance the scale adaptability of the module to input samples and capture different levels of information in the image. At the end of the branch, the dilated convolution is employed. Compared with the ordinary pooling process, the dilated convolution expands the receptive field without reducing the image resolution, which is beneficial to enrich the image features extracted by the module (Yu and Koltun, 2016). After the dilated convolution processing, the four branches are spliced in the channel dimension and processed by 1 × 1 convolution, then merged with the data of another branch, and finally output by Relu activation layer.

Following RFB-s modules, the generator network has 4 convolutional layers and 2 deconvolutional layers. Instance normalization is used to normalize the output results of all convolutional layers to speed up the network training process. Except that the activation function of the last convolution layer adopts tanh, all other convolution layers adopt Relu as the activation function.

In this paper, referring to the method of (Noroozi et al., 2017), the global skip connection is used in the generator, which adds the input of the network directly to the output. In this way, the content that generator learns is the residual between the blurred images and the sharp images. It can be denoted:

$$I_S = I_B + I_R \qquad (5)$$

where $I_R$ is the residual of the latent sharp image $I_S$ and blurred image $I_B$. The principle behind is that, since blurred images and sharp images have many similar image features in terms of color and style, compared to letting the network learn the mapping from blurred images to sharp images, only letting the network learn the residual between them can reduce the difficulty of network learning and make the network converge faster during the training process.

### 3.1.2 Discriminator

In the proposed network, the discriminator is responsible for discriminating sharp images and restored images. While continuously improving the discrimination ability, the discriminator feeds back the identification results (adversarial loss) to the generator to guide the generator's learning.

The discriminator in our network is similar to PatchGAN (Isola et al., 2017). Different from the ordinary discriminator, PatchGAN divides the entire image into several patches, and the output discrimination result is a two-dimensional matrix, and each element in the matrix represents the discrimination result of the corresponding patch. Obviously, PatchGAN are able to guide the generator to perform deblurring learning more precisely than the ordinary discriminators (Isola et al., 2017).

### 3.2 Loss Function

In this paper, the loss function includes three components: adversarial loss, feature loss, and L2 loss.

### 3.2.1 Adversarial loss

Since the adversarial loss of the original GAN has many problems such as mode collapse, gradient vanishing/exploding, in order to avoid these problems and make the training process of the network more stable, the Wasserstein distance (Arjovsky et al., 2017) with gradient penalty (Gulrajani et al., 2017) is used to represent the adversarial loss, which can be denoted by

$$L_{adv} = \sum_{n=1}^{N}(D(I_S) - D(G(I_B))) \\ + \lambda_{GP} \cdot E_x[\|\nabla_x D(x)\| - 1]^2 \qquad (6)$$

where $N$ is the number of images, $D$ is the discriminator, $G$ is the generator, and $G(I_B)$ is the restored image. The first item in (5) represents the scoring of the sharp image and the restored image by the discriminator. In the training process, the discriminator $D$ maximizes the adversarial loss to improve the score of the clear image and reducing the score of the restored image, while the generator $G$ minimize the adversarial loss to improve the score of the restored image, as in (Goodfellow et al., 2014). Through the confrontation between them to improve the deblurring performance of the network. The second term of (5) is the gradient penalty term, and $\lambda_{GP}$ is its weight constant.

### 3.2.2 Feature loss

In DeblurGAN (Kupyn et al., 2018), the author employed the content loss to improve the feature similarity between the restored image and the sharp image, which used the features of conv_3-3 layer extracted from VGG19 network. Considering that convolutional neural network has strong image feature extraction abilities, and the image feature levels extracted from the shallow and deep layers of the network are different, the shallow layers of the network is good at extracting simple color, edge and texture feature, and the deep layers of the network is good at extracting high-level semantic feature (Zeiler and Fergus, 2014). Therefore, only using a certain layer of the deep convolutional neural network to construct the loss function of the feature does not make full use of the feature extraction ability of the convolutional neural network.

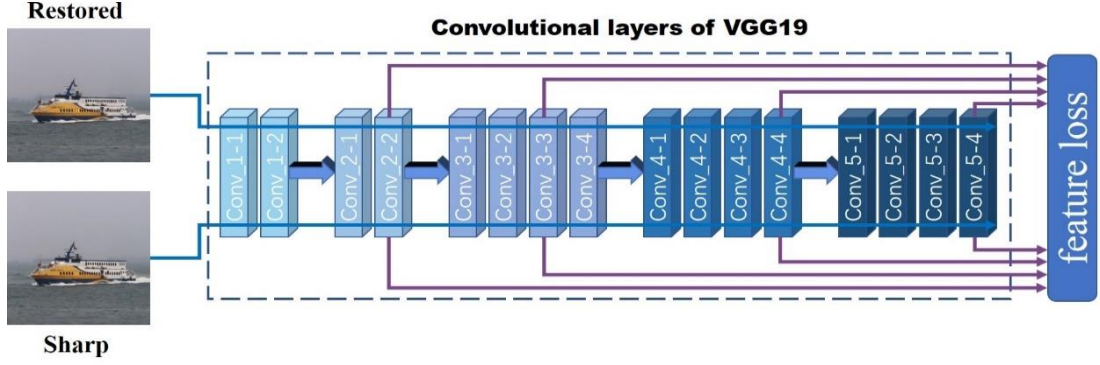

Figure 3: The feature loss fuses the different layers of features extracted from VGG19 network.

In this paper, we proposed the feature loss based on deep convolutional neural network, which uses the feature maps of the shallow and deep layers of the network to fuse different levels of image features to help the network better learn motion deblurring. This is another major innovation of the SharpGAN. In figure 3, we take the VGG19 network as an example to introduce the implementation process of the feature loss.

The VGG19 network has a total of 16 convolutional layers and 3 fully connected layers. When calculating the feature loss, the sharp image and restored image are input to the VGG19 network respectively to obtain their feature maps at different layers, and the mean square error of these feature maps is calculated to reflect the feature difference between the sharp image and the restored image. In order to effectively cover the image features of different levels extracted by the network, conv_2-2, conv_3-3, conv_4-4 and conv_5-4 layers are selected to construct feature loss.

The features loss function is defined by:

$$L_{X_{i,j}} = \frac{1}{W_{i,j} H_{i,j}} \sum_{x=1}^{W_{i,j}} \sum_{y=1}^{H_{i,j}} (\phi_{i,j}(I_S)_{x,y} - \phi_{i,j}(G(I_B))_{x,y})^2 \quad (7)$$

where $\phi_{i,j}$ is the feature map obtained by the j-th convolution (after activation) before the i-th maxpooling layer within the VGG19 network, $W_{i,j}$ and $H_{i,j}$ are the width and height of the feature maps.

3.2.3 *L2 loss*

In order to minimize the pixel error between the restored image and the sharp image, we add the L2 loss to the loss function as in (Nah et al., 2017). The L2 loss function is as follow:

$$L_2 = \sum_{n=1}^{N} (I_S - G(I_B))^2 \quad (8)$$

Combining adversarial loss, feature loss, and L2 loss can get the total loss of our network:

$$\arg\min_G \arg\max_D L(D,G) = L_{GAN} + \lambda_X \cdot L_X - \lambda_2 \cdot L_2 \quad (9)$$

where $\lambda_X$ is the weight constant of feature loss, and $\lambda_2$ is the weight constant of L2 loss.

## 4. Training

In this paper, the GOPRO (Nah et al., 2017) and SMD datasets (Prasad et al., 2017) are exploited to reveal the performance of the proposed SharpGAN.

The GOPRO dataset (Nah et al., 2017), which is widely used in the community of deblurring, employs high frame rate cameras to record videos of different dynamic scenes. The average value of multiple consecutive frames in the video is used to simulate realistic blurred images. There are 2013 and 1111 of blurred and sharp image pairs in the training set and the test set, respectively.

Table 1. Components of each model

| Name | Components |
|---|---|
| Base Model | RFBNet + Content loss based on the conv_3-3 layer of the VGG19 network |
| Base Model+ | RFBNet + Feature loss with multi-level image features |
| SharpGAN | RFBNet + Feature loss with multi-level image features + multi-scale training |

The SMD dataset (Prasad et al., 2017) is taken in the sea area near Singapore and is divided into two parts: on-shore video and on-board video. We randomly selecte 2000 images from the SMD dataset, of which 1500 are employed as the training set and 500 are employed as the test set. Then, the motion blur is added to the sharp SMD dataset. To better fit the actual motion blur problem, we set multiple blur sizes from 16 pixels to 40 pixels, and the blur angle is randomly sampled from 0 ° to 360 ° with uniform distribution. Finally, we add Gaussian white noise with standard deviation σ=0.01 to each blurred image. After the above process, the total images in the training set and the test set are 3000 and 1000, respectively.

The training has a total of 500 epochs for each dataset. The learning rate of the first 250 epochs is set to $10^{-4}$, and then the last 250 epochs decay linearly to $10^{-5}$. The batchsize in the training stage is set to 1. In order to balance the proportion of each loss in the total loss, we set the weight constant $\lambda_{GP}$ of gradient penalty term to 10, the weight constant $\lambda_2$ of L2 loss to $10^6$, and the weight constant $\lambda_X$ of feature loss to 1. Through experiments, we determined that the conv2-2, conv3-3, conv4-4, and conv5-4 layers of the VGG19 network (Simonyan and Zisserman, 2015) accounted for 0.2, 0.4, 0.2, 0.2 in the feature loss. So the final feature loss $L_X$ in the experiment be denoted:

$$L_X = 0.2 L_{X_{2,2}} + 0.4 L_{X_{3,3}} + 0.2 L_{X_{4,4}} + 0.2 L_{X_{5,4}} \quad (10)$$

In the training stage, we proposed a multi-scale training schme. In each step of the training process, we randomly select a scale from 256×256, 384×384, 512×512, and 640×640 to crop the images used for network training to adapt to different scales of motion blur. After that, the randomly mirror or geometrically flip is used to the cropped images to augment the training set.

In order to reveal the improvement of network performance more clearly, we successively employ the RFBNet module, the feature loss with multi-level image features and the multi-scale training method to the network. The specific information of each model in the training process is shown in Table 1, where the trainings of Base Model and Base Model+ used image patches with a size of 512×512.

## 5. Experimental Results

We implement our model with TensorFlow, and all the following experiments are performed on a single NVIDIA GeForce GTX 2080Ti GPU. The PSNR and SSIM are selected as the evaluation criteria of the experiments.

*5.1 GOPRO Dataset*

In the field of motion deblurring, the GOPRO dataset (Nah et al., 2017) is one of the most popular datasets. In order to verify the deblurring performance of SharpGAN, we first compared it with the existing algorithms on the GOPRO dataset. The experimental results are shown in Table 2. It can be seen from Table 2, SharpGAN has obvious advantages in PSNR and real-time performance of deblurring, and has a slight gap in SSIM compared with the method of (Nah et al., 2017).

The deblurred images on the GOPRO dataset (Nah et al., 2017) are shown in Figure 4.

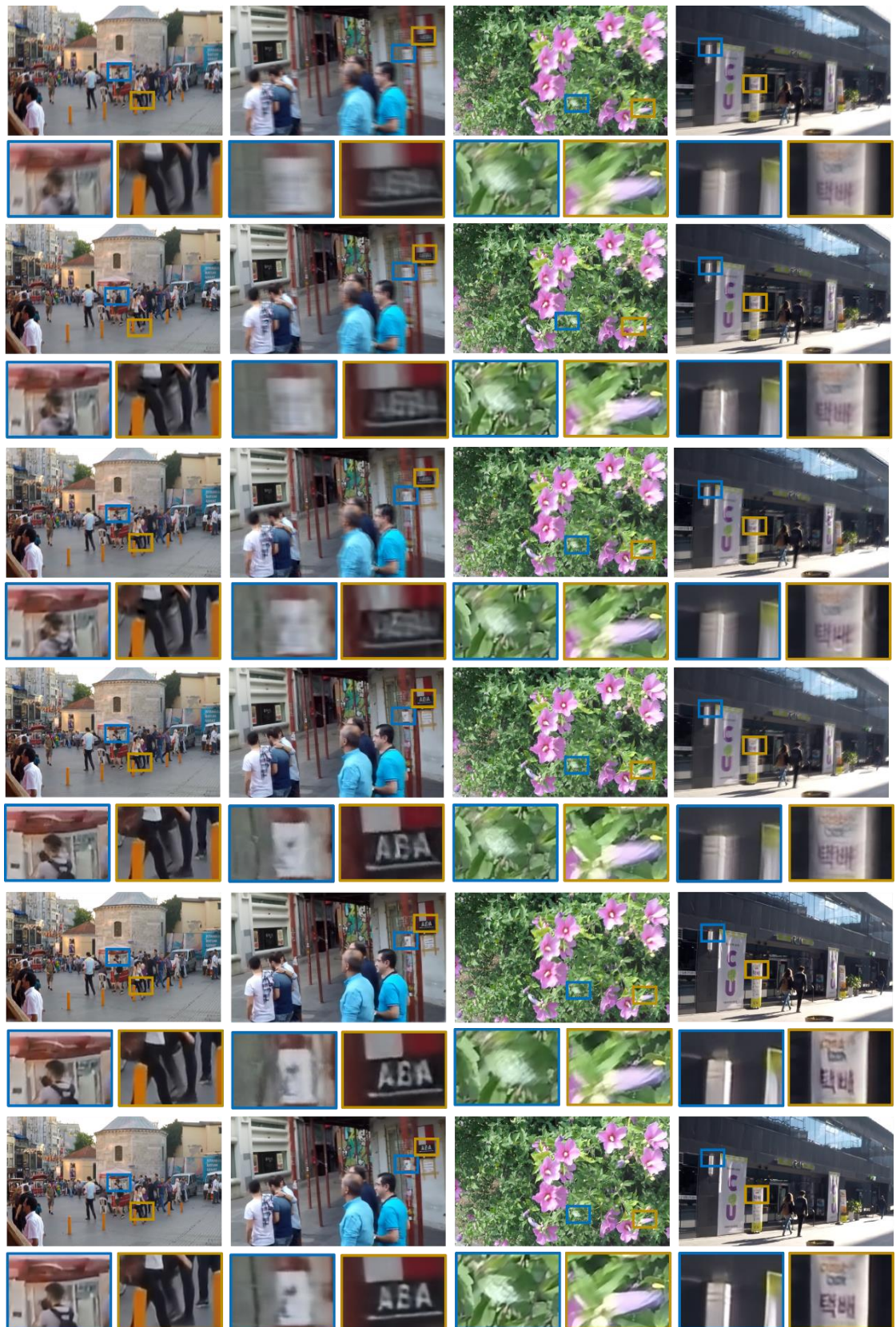

Figure 4: The deblurring results of the GOPRO dataset (Nah et al., 2017) . From top to bottom: the blurred image, the result of (Sun et al., 2015), (Gong et al., 2017), (Nah et al., 2017), (Kupyn et al., 2019) and the result of SharpGAN.

Table 2. The experimental results of different methods on the GOPRO dataset.

| Method | PSNR | SSIM | Time |
| --- | --- | --- | --- |
| (Whyte et al., 2010) | 23.80 | 0.836 | 30min |
| (Sun et al., 2015) | 24.64 | 0.843 | 18min |
| (Gong et al., 2017) | 26.06 | 0.863 | 18min |
| (Nah et al., 2017) | 29.23 | **0.916** | 1.13s |
| (Kupyn et al., 2019) | 27.63 | 0.855 | 0.18s |
| SharpGAN | **29.62** | 0.897 | **0.17s** |

Table 3. The experimental results of different methods on the SMD dataset

| Method | PSNR | SSIM | Time |
| --- | --- | --- | --- |
| (Xu et al., 2013) | 28.59 | 0.792 | 18.68s |
| (Yan et al., 2017) | 27.24 | 0.637 | 10min |
| (Nah et al., 2017) | 30.88 | **0.856** | 2.11s |
| (Kupyn et al., 2018) | 25.75 | 0.596 | 0.57s |
| (Kupyn et al., 2019) | 29.05 | 0.799 | 0.39s |
| Base Model | 31.71 | 0.832 | **0.37s** |
| Base Model+ | 31.84 | 0.835 | **0.37s** |
| SharpGAN | **31.90** | 0.837 | **0.37s** |

It can be seen from Figure 4, SharpGAN has the good deblurring effect in complex dynamic scenes. Compared with the contrast methods, it has better ability to restore the texture details of the objects in the blurred images, and the restored images have the best visual perception effect.

*5.2 SMD Dataset*

To verify the effectiveness of SharpGAN in removing motion blur of real sea images, we compared it with the existing deblurring methods on the SMD dataset (Prasad et al., 2017). The experimental results are shown in Table 3.

It can be seen from Table 3, the PSNR of Base Model is higher than that of all the methods in the comparison. By introducing the feature loss with multi-level image features and multi-scale training methods into the network, both PSNR and SSIM are gradually improved. Though the SSIM of SharpGAN has a very small gap with the method of (Nah et al., 2017), the PSNR has outstanding advantages compared to all the methods. While maintaining the best visual quality of the deblurred images, the proposed method also significantly shortens the deblurring time. The deblurred images on the SMD dataset (Prasad et al., 2017) are shown in Figure 5.

It can be seen from Figure 5, comparing to the existing methods, SharpGAN can more effectively improve the clarity of the blurred image, and better restore the texture detail information of the ship's mast and superstructure. In addition, in order to verify the improvement of the object detection effect of the deblurring methods, we use the pretrained SSD ship object detection model of (Prasad et al., 2017) to detect the blurred images, sharp images and restored images of the different methods. Some of the randomly selected test results are shown in Figure 6.

From Figure 6, we can see that, comparing to the existing algorithms, the SharpGAN can significantly improve the object detection performance of the blurred real sea image, and it can better reconstruct the contour features of the small objects, so that the detection results of the restored images are able to approach to that of the sharp images.

Figure 5: The deblurring results of SMD dataset (Prasad et al., 2017). From top to bottom: Blurred images, results of (Xu et al., 2013), (Yan et al., 2017), (Nah et al., 2017), (Kupyn et al., 2018), (Kupyn et al., 2019) and the results of SharpGAN.

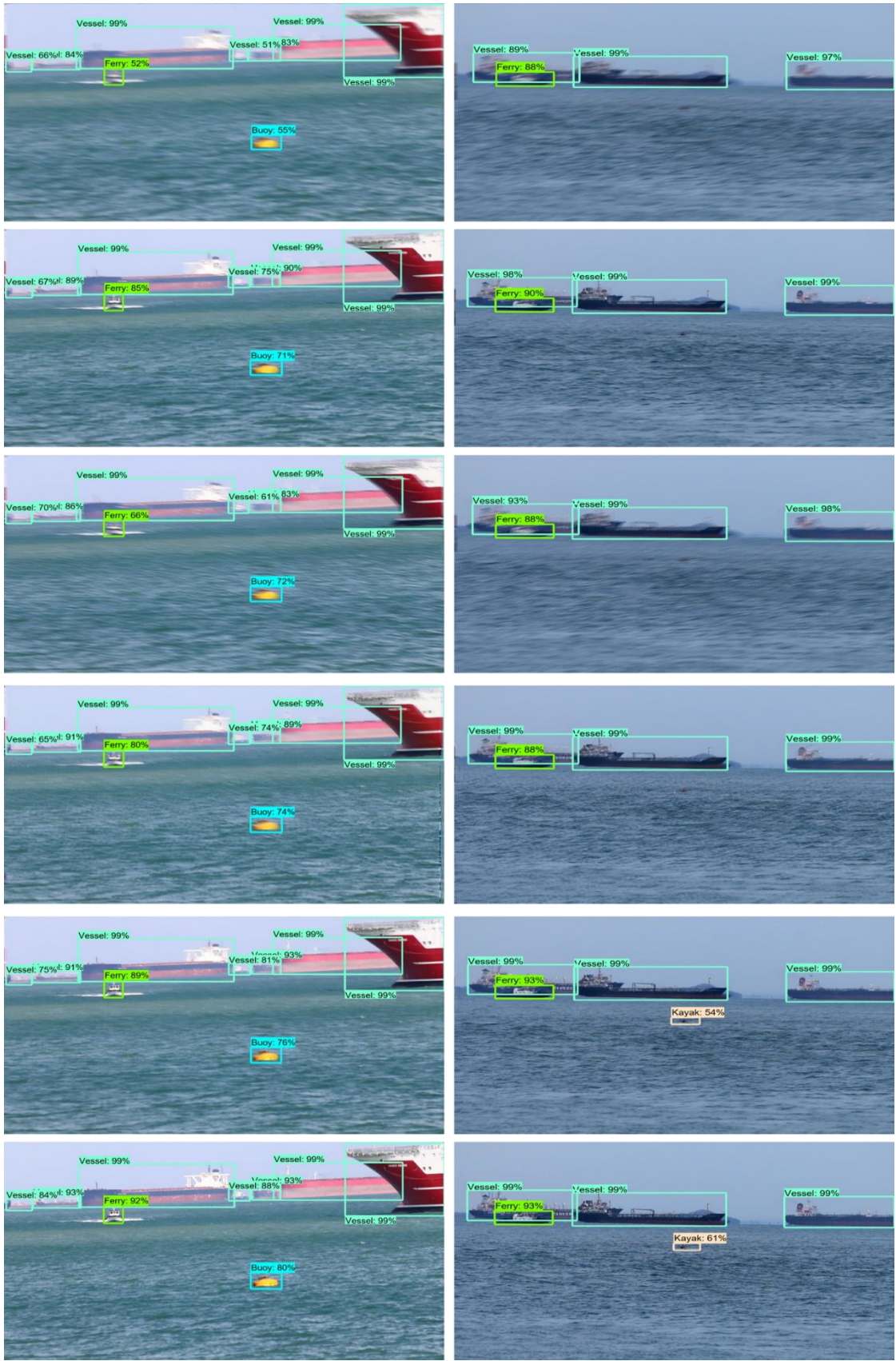

Figure 6: The object detection results on the SMD dataset. From top to bottom: results of blurred images, results of (Nah et al., 2017), (Kupyn et al., 2018), (Kupyn et al., 2019), SharpGAN and the results of sharp images.

**6. Conclusion**

In this paper, we proposed SharpGAN, a new deblurring method based on the generation adversarial networks. Unlike previous studies, the RFBNet module (Liu et al., 2018) was introduced into the deblurring network, and the feature loss was proposed to fusion multi-level image features. The experimental results showed that, compared with the existing methods, SharpGAN had obvious advantages in improving the qualitative and quantitative criteria of the visual perception of the deblurred images, and can also improve the object detection effect of blurred real sea images.